\documentclass[runningheads]{llncs}
\usepackage[T1]{fontenc}
\usepackage{graphicx}
\usepackage{soul,color,ulem, wrapfig}
\usepackage{booktabs}
\usepackage{algorithm}
\usepackage[noend]{algpseudocode}

\begin{document}
\title{Faster Convergence with Lexicase Selection in Tree-based Automated Machine Learning}
\author{Nicholas Matsumoto \and
Anil Kumar Saini \and Pedro Ribeiro \and Hyunjun Choi \and Alena Orlenko \and Leo-Pekka Lyytikäinen \and Jari O Laurikka \and Terho Lehtimäki \and Sandra Batista \and Jason H. Moore}

\authorrunning{Matsumoto et al.}
\titlerunning{Faster Convergence with Lexicase Selection}
\institute{Cedars-Sinai Medical Center, Los Angeles, CA 90048, USA \\
\email{\{nicholas.matsumoto,anil.saini,pedro.ribeiro,hyunjun.choi,alena.orlenko,\\ sandra.batista\}@cshs.org, jason.moore@csmc.edu} \\
Tampere University\\
\email{leo-pekka.lyytikainen@tuni.fi, terho.lehtimaki@uta.fi}\\
Sydänsairaala Hospital\\
\email{jari.laurikka@sydansairaala.fi}
}
\maketitle              % typeset the header of the contribution

\begin{abstract}
In many evolutionary computation systems, parent selection methods can affect, among other things, convergence to a solution. In this paper, we present a study comparing the role of two commonly used parent selection methods in evolving machine learning pipelines in an automated machine learning system called Tree-based Pipeline Optimization Tool (TPOT). Specifically, we demonstrate, using experiments on multiple datasets, that lexicase selection leads to significantly faster convergence as compared to NSGA-II in TPOT. We also compare the exploration of parts of the search space by these selection methods using a trie data structure that contains information about the pipelines explored in a particular run. 

\keywords{Parent Selection  \and NSGA-II \and Lexicase \and Convergence \and Trie.}
\end{abstract}

\section{Introduction}
In evolutionary computation (EC) systems, just like many machine learning (ML) algorithms such as neural networks, the time and resources required to converge to an acceptable solution are important. For tasks such as classification and  regression, faster convergence to an acceptable solution may form the basis of early stopping criteria in resource-constrained environments. However, even the notion of convergence for EC systems requires careful specification of what convergence means and how it will be measured. This also requires careful consideration of effects of specific aspects of the implementations of EC systems.

Parent selection methods are used in EC systems to select parents from the current generation that are used to produce the next generation. These methods differ in the way they use the fitness of the individuals on different objectives. Consequently, parent selection methods can affect various properties of the evolving population. In this paper, we conduct a case study on the effect of parent selection algorithms on convergence in Tree-based Pipeline Optimization Tool (TPOT)~\cite{tpot2016}, an automated machine learning system that uses genetic programming (GP) to evolve machine learning pipelines for classification and regression tasks.  Specifically, through experiments, we demonstrate that using lexicase selection, as compared to the default selection algorithm, Non-dominated Sorting Genetic Algorithm II (NSGA-II), leads to faster convergence without loss of accuracy on the holdout set. We use a variant of lexicase selection called automatic $\epsilon$-lexicase selection that is used when the objectives are real-valued. We define convergence as the generation when the best model in the population (based on cross-validation accuracy on the training data) attains at least 99\% of the cross-validation accuracy of the best model in the final generation of that GP run.

For our experiments we use synthetic and real clinical datasets. The DIverse and GENerative ML Benchmark (DIGEN)~\cite{orzechowski2021generative}  provides 40 synthetic binary classification datasets that were designed to produce a diverse distribution of performance scores for different popular machine learning algorithms. The other one is the Angiography and Genes Study (ANGES) dataset that came from the study conducted at Tampere University Hospital, Finland. The study includes 925 patients and provides their clinical data, measured coronary arteries angiography, and metabolic profiling. 

In our experiments with the DIGEN and ANGES datasets, we found that, on average, TPOT with lexicase selection converges multiple generations earlier compared to TPOT with NSGA-II, without sacrificing on the holdout set accuracy or changing the number of machine learning methods used in the best models of the final generations. 

We also looked at the exploratory behavior of TPOT with the selection methods using an \textit{exploration trie} that visualizes sequences of machine learning operators in pipelines TPOT explores. While the tries for both selection methods had similar depth, we found that, with NSGA-II, tries for TPOT are larger with more balanced branching. Lexicase selection, on the other hand, leads to smaller tries with more branching in certain directions. The selection methods, therefore, exhibit different behaviors when it comes to prefixes of sequences of machine learning operators used in pipelines. NSGA-II tends to explore pipelines with many different prefixes, but lexicase tends to focus on a few prefixes and explore around them.

Various sections in this paper are organized as follows. We start with related work in the GP literature in Section \ref{sec:related}. Then after describing the TPOT system and the selection methods studied in this paper in Section
\ref{sec:methods}, we describe the experimental design and the datasets used in Section \ref{sec:experiments}. The results are presented in Section \ref{sec:results} and discussed in Section \ref{sec:discussion}.

\section{Related Work}
\label{sec:related}

\textbf{Convergence.} In computer science literature, the term convergence is defined in a variety of contexts. In neural networks, for example, parameters of the networks are said to be converging when the corresponding weights are not changing on account of gradient descent or other updates~\cite{oh2020convergence}. In genetic programming, convergence is sometimes defined in terms of diversity, i.e., if the proportion of unique individuals in the population is very low, the population is said to have converged~\cite{langdon2022genetic,ciesielski2002prevention}. The term `uniqueness' may itself be defined in terms of the individuals which have the same outputs, same size, or any other characteristics. In this study, however, we define convergence in a somewhat related way as the generation after which there is no visible change in the performance of the best individual in the population.

\vspace{\baselineskip}

\noindent \textbf{Comparison among selection methods. }Multiple works analyze the effect of parent selection methods on certain properties of evolving populations such as their diversity (e.g., number of unique genotypes) and modularity. In Metevier et al. \cite{metevier2019lexicase}, the authors compared lexicase selection, tournament selection, and fitness proportionate selection on the success rate, the number of generations used to find a solution, and structural diversity. In Saini and Spector~\cite{saini2021relationships}, the authors demonstrated that lexicase selection, compared to other selection methods, leads to a significantly greater number of individuals with looping instructions in the evolving population. 

In multi-objective and many-objective optimization settings, lexicase selection and its variants have been compared to NSGA-II and other methods. In evolving gaits in quadrupedal animats, NSGA-II significantly outperforms lexicase selection when distance traveled, efficiency, and vertical torso movement were used as objectives during evolution~\cite{moore2016comparison}.
When experimenting on many-objective optimization problems like DTLZ~\cite{deb2005scalable}, lexicase selection outperforms NSGA-II, especially when the number of objectives is more than 5~\cite{la2018analysis}.

\vspace{\baselineskip}

\noindent \textbf{Analyzing GP runs.} Various works~\cite{mcphee2018detailed,burlacu2013visualization} try to summarize information about GP runs in a graphical form. For example, McPhee et al. \cite{mcphee2018detailed} presents a way to record information about genetic ancestry in a graph database, which basically means recording the individuals in a particular run which contributed to the material in the final solution. In this paper, we introduce a different method of recording information in exploration tries.

\section{Methods}
\label{sec:methods}
\subsection{Review of TPOT}
TPOT is a tree-based genetic programming system implemented in Python that searches for machine learning pipelines for classification and regression tasks. A given evolving individual is a machine learning pipeline including its methods and hyperparameters represented as a tree. Within pipelines, different machine learning methods or \textit{operators} such as Random Forest, PCA, and RFE, may be applied in composition on the data. TPOT uses the Distributed Evolutionary Algorithm in Python (DEAP) \cite{DEAP_JMLR2012} framework to evolve pipelines by using the variation operators and selection methods as implemented in DEAP. Using DEAP's use of primitives and terminals, we describe primitives in TPOT as single machine learning modules or operators and the terminals as their respective hyperparameters. For classification tasks, TPOT currently supports 32 operators with corresponding hyperparameters. 

\subsection{Parent Selection Algorithms}
The procedure to generate individuals for the next generation from the current evaluated individuals is given in Algorithm \ref{alg:select}. From the current population, we generate the same number of individuals through crossover and mutation.
For crossover, we search in the current population for the individuals which share at least one primitive using a function \textit{ChooseEligibleIndividuals()}. Then after performing crossover, we choose the first child. If that child has already appeared as one of the offspring, we discard it, and instead apply mutation on a randomly chosen individual from the population.
Note that \textit{ChooseRandomly()} chooses the individuals uniformly at random. The mutation operation proceeds in the regular fashion.

After applying the variation operators, we input the parent and offspring individuals to one of the selection methods. The description of the different selection methods used in this study is given in the following sections.
\vspace*{-.4cm}
\begin{algorithm}
\caption{Selection Procedure}
\label{alg:select}
\begin{algorithmic}[1]

\Procedure{select}{$curr\_pop, cross\_prob, mut\_prob, method, pop\_size$}       
%\Comment{Generate inds. for the next gen.}
    \State \textit{offspring}=[]
    \While{\textit{size(offspring)} $\not=$ \textit{pop\_size} }
    \State rand = rand()
    \Comment{Generate a random number from [0,1)}
    \If{$rand < cross\_prob$}
        \State $ind1, ind2$ = ChooseEligibleIndividuals(curr\_pop)
        \State $ind3, ind4$ = Crossover(ind1, ind2)
        \If{$ind3$ is a duplicate from \textit{offspring}}
            \State $ind3$ = Mutation(ChooseRandomly(curr\_pop))
        \EndIf
        \State offspring.append($ind3$)
    \Else
        \State $ind1$ = ChooseRandomly(curr\_pop)
        \State $ind2$ = Mutation(ind1)
        \State offspring.append($ind2$)
    \EndIf
    \EndWhile
    \If{method=NSGA2}
        \State parents = NSGA2(curr\_pop + offspring, pop\_size)
    \Else 
         \State parents = $\epsilon$Lex(curr\_pop + offspring, pop\_size)
    \EndIf
    
\EndProcedure
\end{algorithmic}
\end{algorithm}
\vspace*{-.8cm}
\subsubsection{NSGA-II:}
Non-dominated Sorting Genetic Algorithm II (NSGA-II) is a multi-objective evolutionary algorithm that has been widely used for optimization problems with multiple objectives~\cite{deb2002fast}.

For every selection event, as shown in Algorithm \ref{alg:nsga}, we combine the parent and offspring populations and perform non-dominated sorting. The sorting will lead to the combined population getting organized into `fronts', whereby individuals in front 1 dominate\footnote{Individual $i_1$ dominates $i_2$ if  $i_1$ is better than or the same as  $i_2$ on all objectives and strictly better than  $i_2$ on at least one objective.} individuals in front 2, and so on. Consequently, in Algorithm \ref{alg:nsga}, \textit{ParetoFrontsNonDominatedSort()} returns a list of lists of individuals. For all of the individuals in various fronts, we also calculate a metric called `crowding distance' which is a measure of the density of individuals around a particular individual in the objective space (see \cite{deb2002fast} for more details).

Then, we start adding individuals from the fronts to the new population until its size reaches $n$ individuals (with $n$ being the size of population, $pop\_size$). If the size of the first front is more than the population size, we sort the individuals by crowding distance and keep the first $n$ individuals. Otherwise, we add the whole front to the population and look at the second front. If the total individuals in fronts 1 and 2 is more than $n$, we sort the second front by crowding distance and keep the first $(n-size(front 1))$ individuals. Otherwise, we look at the subsequent fronts, and repeat the process as summarized in lines 4-8 in Algorithm \ref{alg:nsga}.

\vspace*{-.4cm}
\begin{algorithm}
\caption{NSGA-II}
\label{alg:nsga}
\begin{algorithmic}[1]
\Procedure{NSGA2}{pop, pop\_size}
    \State $parents = [], i=1$
    \State $fronts$ = ParetoFrontsNonDominatedSort(pop)
    \Comment{list of lists of individuals}
    \While{$size(parents) + size(fronts[i]) \leq pop\_size$}
        \State $parents = parents + fronts[i]$
        \State $i = i+1$
    \EndWhile
    \State sort $fronts[i]$ using crowding distance
    \State $parents = parents + fronts[i][:pop\_size-size(parents)]$
    
\EndProcedure
\end{algorithmic}
\end{algorithm}
\vspace*{-.8cm}
\subsubsection{Lexicase Selection}
Lexicase selection~\cite{helmuth2014solving,metevier2019lexicase} is a parent selection method used in genetic programming and other evolutionary computation techniques. Our implementation of lexicase selection is given in Algorithm \ref{alg:lex}. For every selection event, first, we randomly shuffle the list of objectives on which a given individual is evaluated. Then, the pool of candidates, which initially contains the whole population, is whittled down based on their performance on the objectives: the individuals that perform the best on the first objective are kept in the pool and others are removed. Then from this pool, the individuals that perform the best on the second objective are kept in the pool and others are removed, and so on. The process is repeated until we are left with only one individual in the pool, or, we are out of objectives. In the second case, we randomly choose one of the individuals as the parent.

$\epsilon$-lexicase selection~\cite{la2016epsilon} and automatic $\epsilon$-lexicase selection~\cite{la2016epsilon} are variants of lexicase selection that have been developed for the settings where individuals can have real-valued errors or fitness values, as for example, in symbolic regression. 
In $\epsilon$-lexicase selection, for a given objective, all the individuals within $\epsilon$ of the fitness of the best individual in the current pool are kept in the pool and the rest are removed. The value of $\epsilon$ is a parameter of the algorithm. It is specified by the user and is fixed during the whole process. Automatic $\epsilon$-lexicase selection has the same process as $\epsilon$-lexicase selection, but the value of $\epsilon$ is automatically determined by the algorithm: the median absolute deviation (MAD) of the fitness values of the individuals in the current pool on a selected objective. In other words, $\epsilon_t = median(|x_1 - median(x)|, |x_2 - median(x)|, ... )$, where $x_1, x_2, ...$ are the fitness values of the individuals in the \textit{current pool} on objective $t$.

We ran experiments on both the regular lexicase and the automatic $\epsilon$-lexicase selection methods. However, we include results for automatic $\epsilon$-lexicase (simply called lexicase from here onwards) only since there was no substantial difference between the two in terms of convergence, accuracy on the holdout set, and the number of operators.

% $\epsilon$ = median(|x - median(x)|), where x is the fitness values of the individuals on one of the test cases.

\vspace*{-.4cm}
\begin{algorithm}
\caption{$\epsilon$-Lexicase Selection}
\label{alg:lex}
\begin{algorithmic}[1]
\Procedure{$\epsilon$Lex}{pop, pop\_size}
    \State $parents = []$
    \While{$size(parents) \not= pop\_size$}
        \State $curr\_pool = pop$
        \State $curr\_objectives$ = objectives sorted in a random order
        \For{$obj$ in $curr\_objectives$}
            \State $best\_val$ = best value on $obj$ in $curr\_pool$ 
            \State $\epsilon$ = median absolute deviation of $obj$ values for $curr\_pool$
            \State $curr\_pool$ = inds. from $curr\_pool$ with $obj$ values within $\epsilon$ of $best\_val$
        \EndFor
        \If{$size(curr\_pool \not=1)$}
            \State $parents$ = $parents$ + one individual from $curr\_pool$ chosen randomly
        \Else
            \State $parents$ = $parents$ + $curr\_pool$
       \EndIf
    \EndWhile
\EndProcedure

\end{algorithmic}
\end{algorithm}
\vspace*{-.9cm}

\section{Experimental Set-up}
\label{sec:experiments}
We begin by summarizing the datasets that we used for our experiments in the following subsection. In subsequent subsections, we give implementation details and parameters for our experiments, the metrics used to evaluate convergence, and finally, the construction and metrics of exploration tries used to examine the behavior of the selection algorithms.

\subsection{Datasets}
\subsubsection{DIGEN:}

The DIverse and GENerative ML Benchmark (DIGEN) is a set of 40 synthetic, binary classification data sets \cite{orzechowski2021generative}. Each dataset consists of an 800 sample training set and a 200 sample testing set. There are 10 features independently generated from a Gaussian distribution. The binary target is generated with a unique generative function. The set of 40 generative functions for the datasets were designed to produce a diverse distribution of performance scores and relative ranking for eight popular machine learning algorithms such as Decision Trees and Gradient Boosting.

After running TPOT with both selection methods on all 40 DIGEN sets, we noticed that within the initial population, TPOT achieved a high average balanced accuracy score for 27 DIGEN datasets on the training set, and there was little change in the later generations. For the present study, since we need a sufficient difference to determine the efficacy of each selection algorithm, we will focus on only 13 of the 40 DIGEN data sets which provided at least a 10\% increase from the average balanced accuracy across all pipelines from the first generation to twentieth generation pipelines.
\subsubsection{ANGES:}   
 The Angiography and Genes Study (ANGES) includes data on 925 Finnish subjects with coronary angiography, or specifically, the evaluation of the degree of coronary artery stenosis, and targeted metabolic profiling (for detailed study population description, see \cite{orlenko2020model}). ANGES dataset contains 73 metabolic and clinical features, and one binary outcome for coronary artery disease (CAD) status, where patients were considered cases for the disease when any major coronary artery  is detecting stenosis greater than 50\% and controls otherwise. The dataset was split into training (75 percent) and testing (25 percent) sets prior to the analysis. 

\subsection{Implementation}
As mentioned earlier, we conduct our experiments in an AutoML system called TPOT\footnote{https://github.com/EpistasisLab/exploration-trie-tpot}. Except for the selection method, the implementation of TPOT was kept constant as we test the effects of NSGA-II and lexicase on the evolution of machine learning pipelines.  For both methods, we use the following  objectives: 
\begin{enumerate}
    \item Maximize the balanced accuracy calculated from the 10-fold cross-validation scores on the training data.
    \item Minimize the number of machine learning operators used in a given pipeline.
\end{enumerate}

During crossover on two individual pipelines, one of the shared primitives is chosen randomly and the subtrees at nodes corresponding to the chosen primitive in both individual trees are swapped.

  %During each mutation operation for a randomly chosen individual pipeline with equal probability, a randomly chosen machine learning operator primitive may be inserted into the pipeline, an operator primitive may be removed from the pipeline, a primitive operator may be replaced with a randomly chosen operator or its hyperparameters may be changed. If an operator is chosen to be replaced, but it is the only operator in a pipeline, replacement by another operator, change of hyperparameters or insertion of another operator is applied instead.
 
 For every mutation operation, one of the following mutation strategies are chosen with equal probability:
\begin{enumerate}
    \item \textit{mutInsert} – Insert a randomly chosen primitive into the tree.
   \item \textit{mutShrink} – Remove an a primitive from the tree. If there is only one primitive in the pipeline, the removal will not take place; instead, the mutInsert or mutNodeReplacement technique would be applied with equal probability.
   \item \textit{mutNodeReplacement} – Replace a randomly selected node in the tree; if the node is a primitive one, it is replaced by a new randomly chosen primitive node, otherwise a terminal node is mutated by changing the values of hyperparameters in that node.
\end{enumerate}
 
The rates of crossover and mutation are 0.1 and 0.9, respectively. This means effectively, 10\% of offspring are produced by crossover and the rest using mutation operator (see lines 3-14 in Algorithm \ref{alg:select}). Table ~\ref{tab:gleam_params} summarizes the parameters used for TPOT for each dataset for the experiments.

\begin{table}
\centering
  \caption{Genetic Programming parameters.}
  \label{tab:gleam_params}
  \begin{tabular}{lp{3.2cm}p{3.2cm}}
    \toprule
     Parameter & Values (ANGES data) & Values (DIGEN data)\\
       \midrule
     Population size (initial gen.) & 100 & 80 \\
     Population size (later gen.) & 50 & 40\\
     Number of generations & 100 & 20\\
     Number of runs per selection method & 50 & 40\\
     Mutation operator & mutInsert, mutShrink, mutNodeReplacement & mutInsert, mutShrink, mutNodeReplacement\\
     Mutation rate & 90\% & 90\%\\
     Crossover operator & one-point crossover & one-point crossover\\
     Crossover rate & 10\% & 10\% \\
  \bottomrule
\end{tabular} 
\end{table} 
%\vspace*{-1cm}

\subsection{Evaluating Convergence}
To evaluate convergence for a TPOT run on a single random seed, we first define our `convergence point' to be the first generation at which any individual pipeline in the population reaches  99\% of the best balanced accuracy found in any individual pipeline in the final generation of that GP run. Therefore, we will have a convergence point for each run launched for a particular data set and selection method combination. Note that we use only one of the objectives while defining convergence, since in most settings, accuracy on the training set is considered a primary objective with objectives such as size of the model considered secondary. Using only one objective while defining convergence does not change the basic behaviour of the algorithms studied here.

We record the balanced accuracy and the number of operators for each individual pipeline in each generation of a TPOT run. Then using the TPOT runs as our experimental samples, we test for any statistically significant difference in the convergence points, the balanced accuracy on the holdout set, and the number of operators in pipelines at the convergence points for the selection methods. The non-parametric Mann-Whitney-U test is used for these comparisons.  Another way we compare the accuracy of the best performing classifiers from each selection method is by constructing binomial confidence intervals for the balanced accuracy of the classifier on the holdout set \cite{dietterich,snedecor}. 

%This is done using t-tests for pairwise comparison of the TPOT selection variants for the following metrics: $a$, maximum balanced accuracy, $g$, stopping time convergence generation, and $o$, number of operators in the pipeline in stopping time convergence generation.

%Let $\mu$ be the mean of a metric and $v_1$ and $v_2$ be any two variants in the set of NGSA-II, lexicase, and $\epsilon$-lexicase.  The testing considers the following null hypotheses:
%\begin{enumerate}
 %   \item $\mu_{a_{v_1}} = \mu_{a_{v_2}}$
  %  \item $\mu_{g_{v_1}} = \mu_{g_{v_2}}$
   % \item $\mu_{o_{v_1}} = \mu_{o_{v_2}}$
%\end{enumerate}
%It is only necessary to have 3 tests, one for each pair of variants, per hypothesis for each data set. We investigate whether the stopping time convergence generation for lexicase variants is significantly sooner than for NGSA-II while the best accuracy is at least as good for lexicase compared to NGSA-II using at most as many operators in lexicase variants as used in NGSA-II.

\subsection{Exploration of Pipelines}

The convergence to a solution in genetic programming is often affected by the extent of the search space being explored. Therefore, we implement an \textit{exploration trie} to investigate the explored search subspace in TPOT. The root node of the trie represents an empty sequence and all other nodes represent\st{s} a machine learning operator. A path from the root to a node in the trie represents a sequence of machine learning operators explored during the run, and as with more general tries, any sequences that share the same prefix of operators will share the same path from the root in the trie. The trie represents the space of possible sequences of machine learning operators considered during the TPOT run and may not correspond directly to the evaluated pipelines with hyperparameters. Therefore, we are using trie graphs only to \textit{compare} the exploration capabilities of the selection methods, instead of summarizing the pipelines explored by them.

\begin{figure}[h]
 \fbox{\includegraphics[width=6cm, height=4cm]{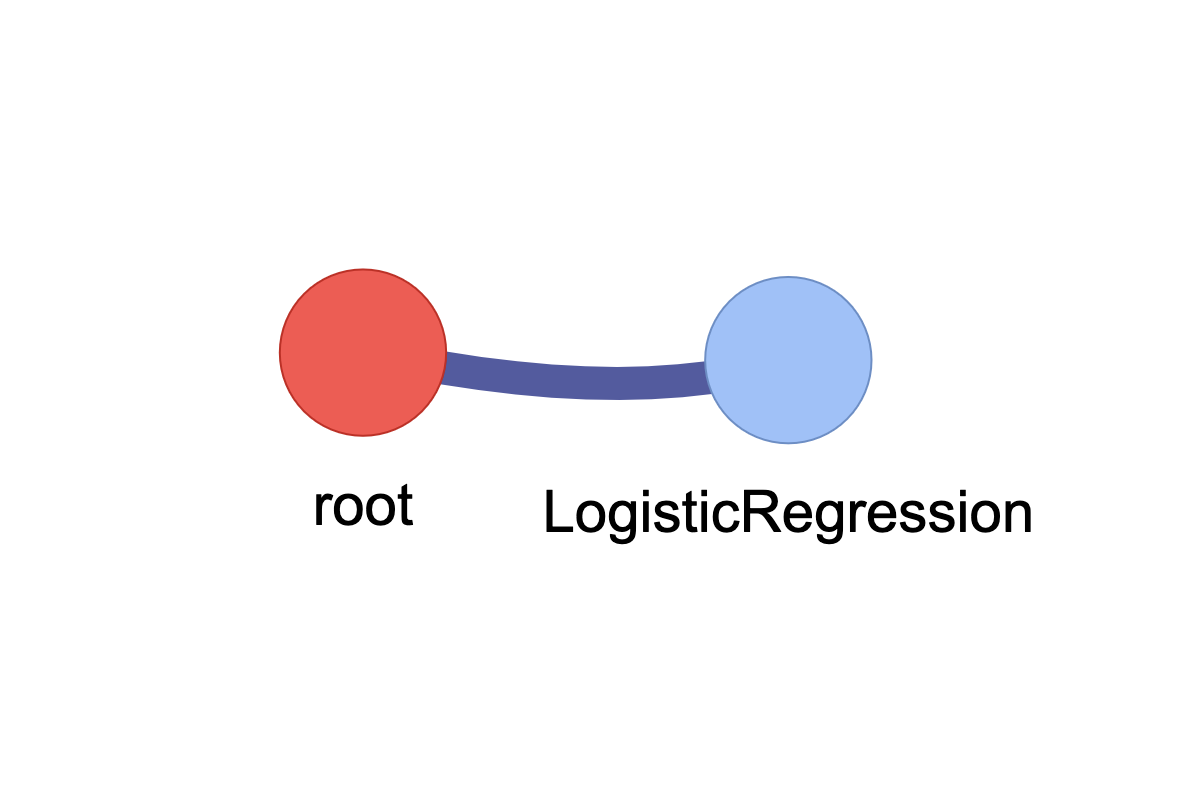}}
 \fbox{\includegraphics[width=6cm, height=4cm]{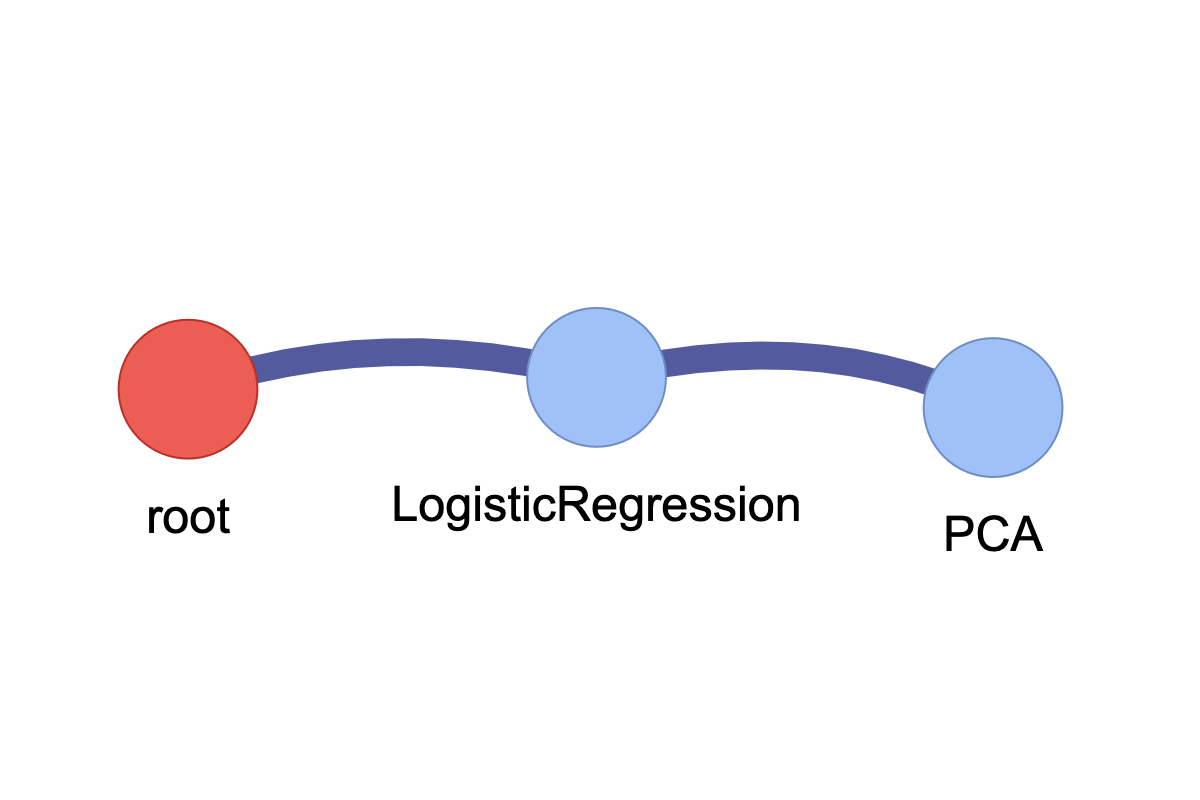}}
 \fbox{\includegraphics[width=6cm, height=4cm]{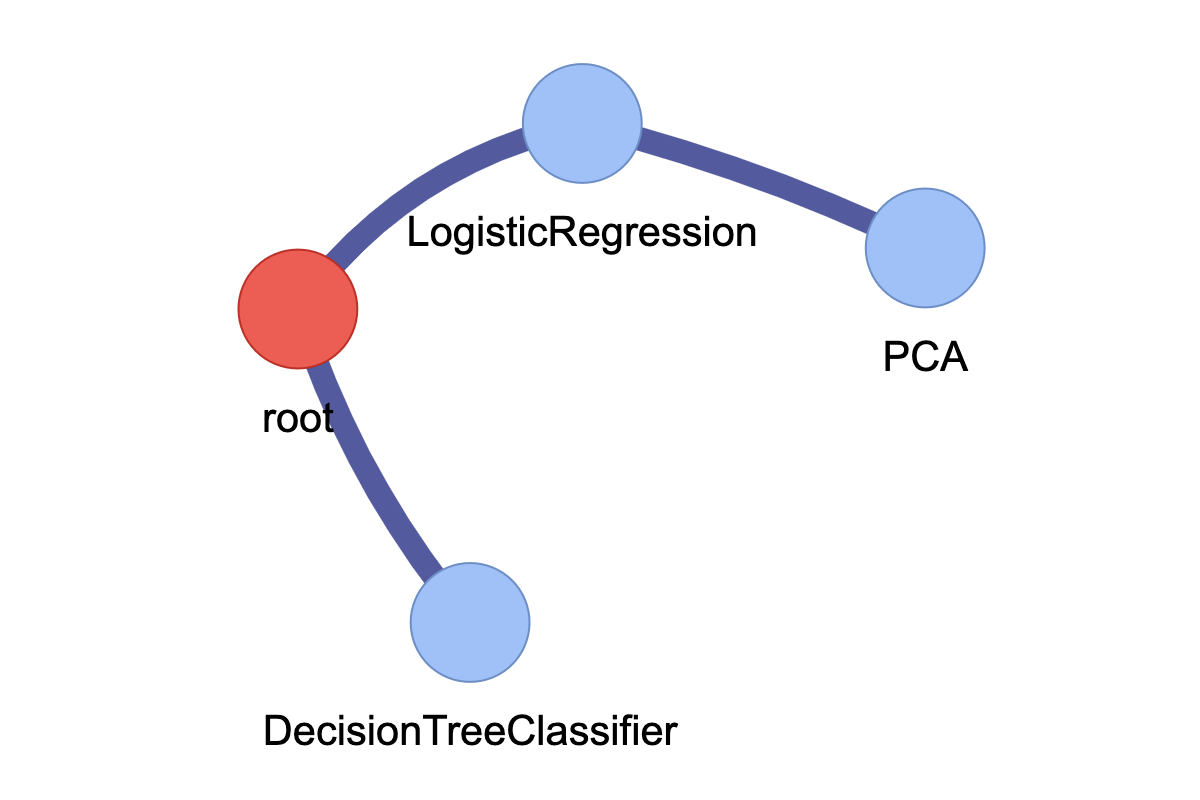}}
 \fbox{\includegraphics[width=6cm, height=4cm]{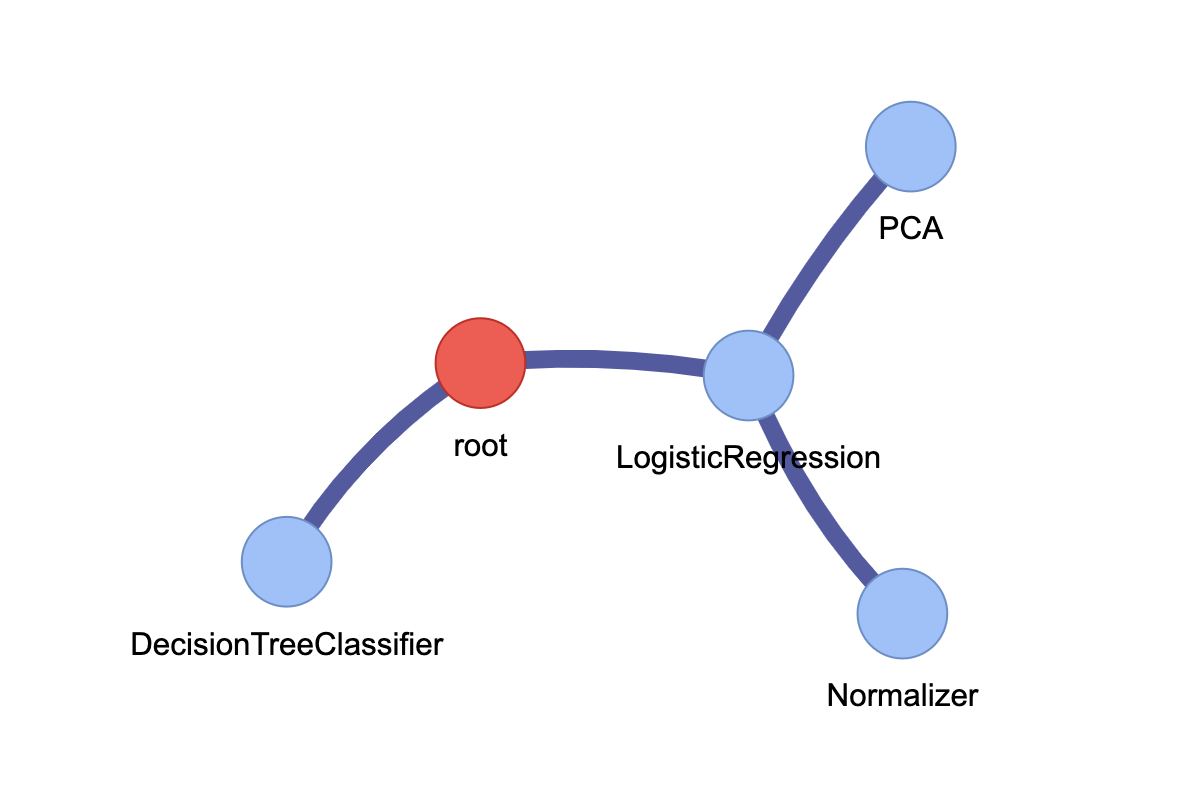}}

 \caption{Sequences inserted in order and the resulting trie graphs after each insertion: LogisticRegression(X), LogisticRegression(PCA(X)), DecisionTreeClassifier(X), LogisticRegression(Normalizer(X)).The root node is shown in red.}
    \label{fig:trie_bldg}
\end{figure}

Let us consider an example exploration trie. In Figure~\ref{fig:trie_bldg}, we trace the construction of the TPOT trie for a TPOT run that has explored the following set of sequences: \{\textit{LogisticRegression(X), LogisticRegression(PCA(X)), DecisionTreeClassifier(X), LogisticRegression(Normalizer(X))} \}. First, starting from an empty sequence representing a root node, TPOT explores the sequence consisting of only logistic regression on the data, $X$. When the second sequence applies PCA to the data before logistic regression is explored, a node for PCA is added to the trie following logistic regression. When TPOT explores the decision tree classifier on the data, a new branch is added to the trie directly from the root. Finally, when TPOT explores a sequence that first normalizes the data and applies logistic regression, a node for a normalizer is added to the trie following logistic regression.

We use graph metrics on the trie such as nodal global efficiency, leaf-to-node ratio, and the number of unique nodes to better understand the evolutionary selective pressure that lexicase and NSGA-II provide.  
\begin{enumerate}
    \item \textbf{Nodal Global Efficiency:} To get a measure of the length of the unique sequences explored, we use the metric called the nodal global efficiency, $g$. It is the average inverse shortest distances between a given node $i$ and all other nodes in the trie. If $T$ is the set of nodes in trie excluding the node $i$, $n$ is the number of nodes in the trie, and $d_{ij}$ is the shortest distance between node $i$ and $j$, the nodal global efficiency~\cite{globalefficiency} for node $i$ can be defined as: $g_i  = \frac{1}{(n-1)} \sum_{j \in T}\frac{1}{d_{ij}} $. In this work, we use nodal global efficiency for the root node (simply called `nodal global efficiency' in this paper). This metric has a value between 0 and 1, where 1 indicates the node is directly connected to all other nodes and a value closer to 0 indicates longer average shortest paths to other nodes (0 would mean the node is disconnected from all other nodes). We calculated the root nodal global efficiencies by constructing the adjacency matrices for tries assuming uniform weights on edges and using Dijkstra's algorithm
    \cite{globalefficiency}.
    \item \textbf{Number of trie nodes:} The number of total nodes excluding the root in the exploration trie is the total number of machine learning operators explored in sequences.
    \item \textbf{Leaf-to-node ratio:} The leaf nodes represent the ends of branches in the trie or the endpoints of the sequences of operators. Accordingly, the leaf-to-node ratio measures the proportion of branches in the trie relative to the total number of nodes. The leaf-to-node ratio is between 0 and 1. The maximum value of leaf-to-node ratio means the root node is directly connected to all other nodes by single edges whereas the value is minimum if there is one linear sequence of all nodes. The leaf-to-node ratio captures the notion of branching in the trie.
\end{enumerate}

\section{Results}
\label{sec:results}
\begin{figure}[h]
\centering
\begin{tabular}{c|c}
     \fbox{\includegraphics[width=5cm]{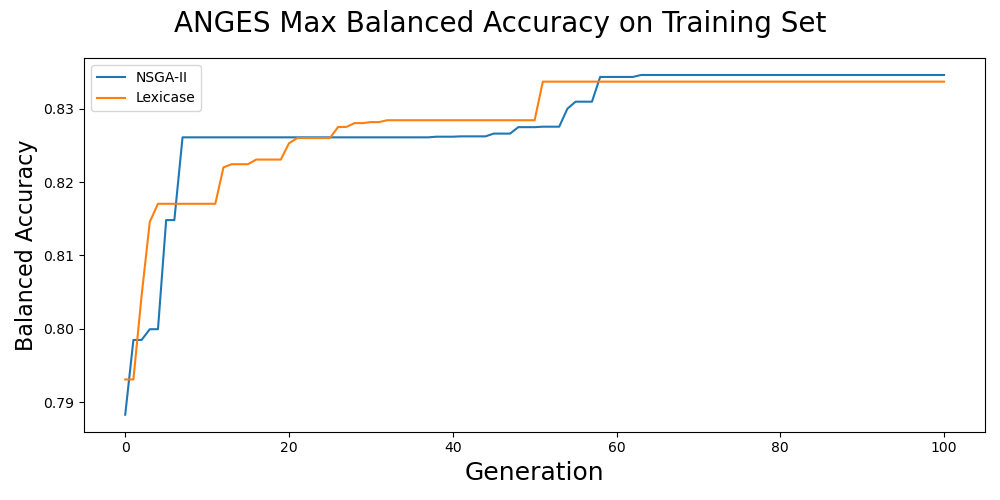}} &  \fbox{\includegraphics[width=5cm]{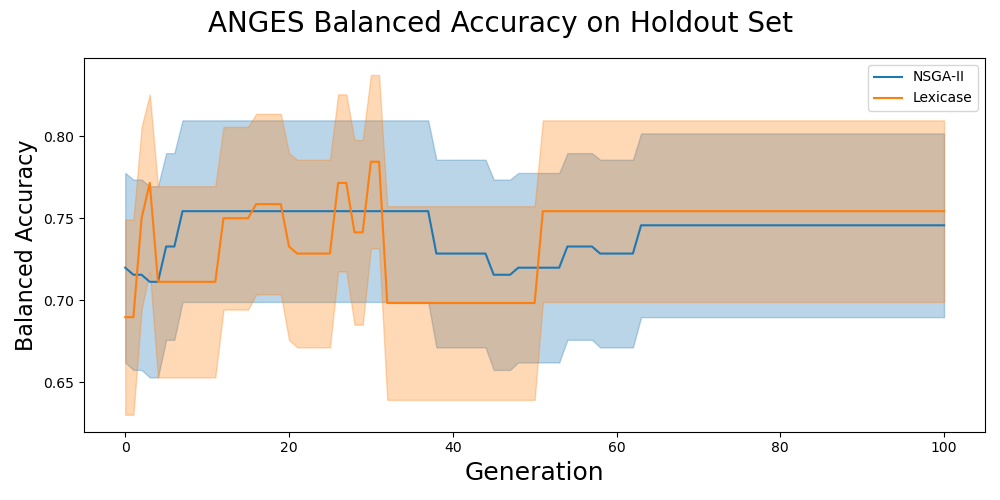}}  \\
     \fbox{\includegraphics[width=5cm]{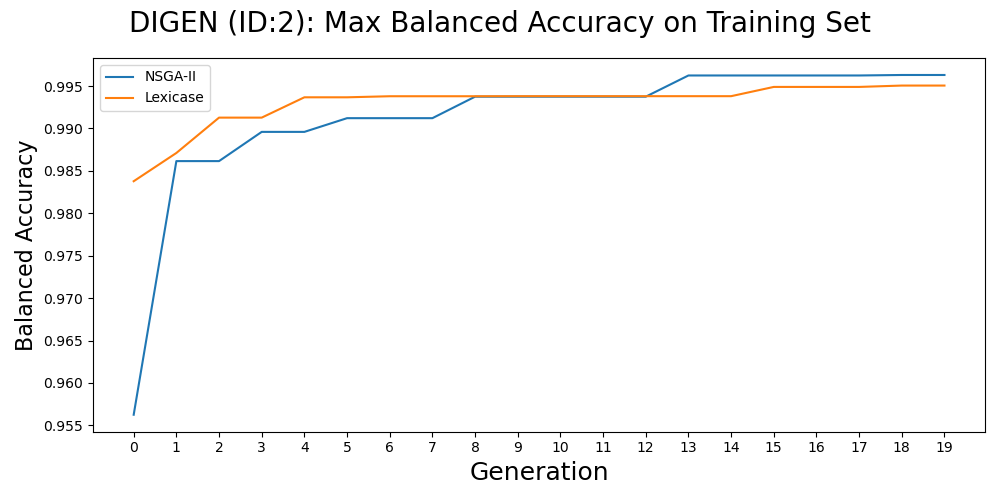}}  & \fbox{\includegraphics[width=5cm]{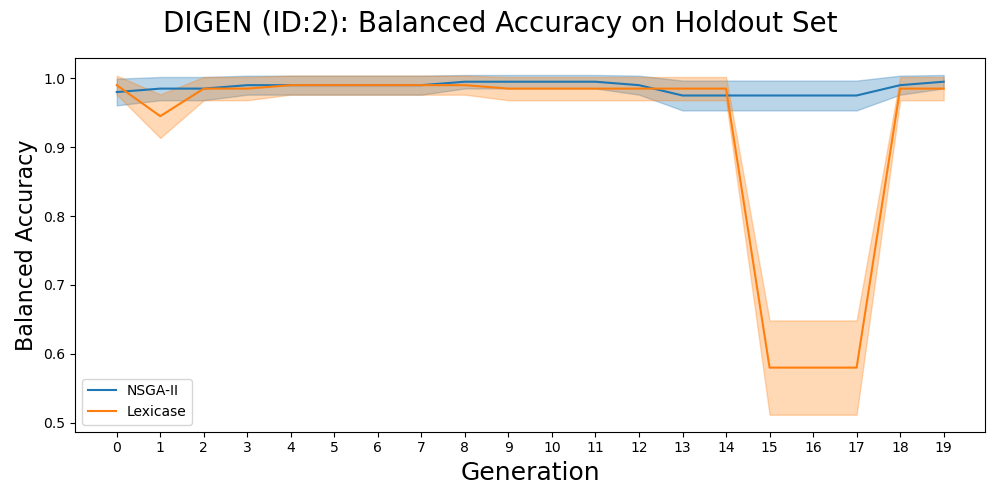}}   \\
    \fbox{\includegraphics[width=5cm]{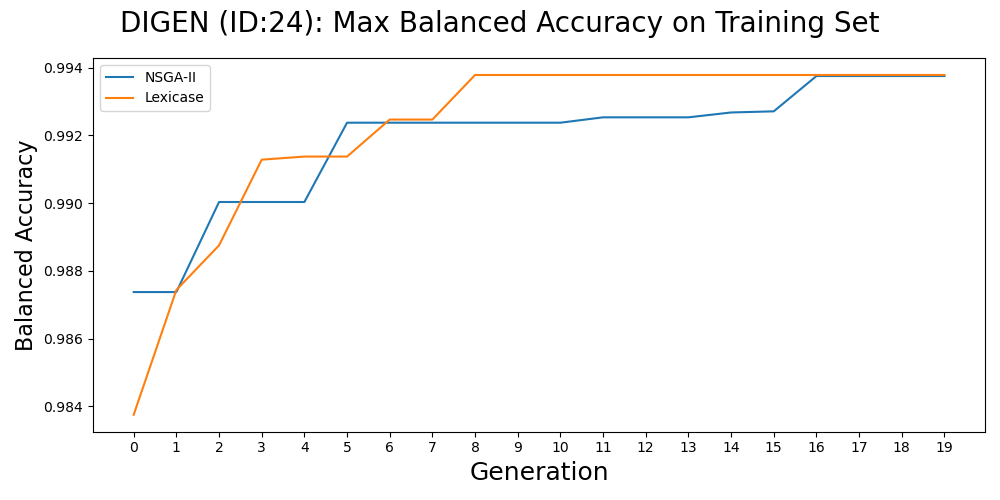}} &  \fbox{\includegraphics[width=5cm]{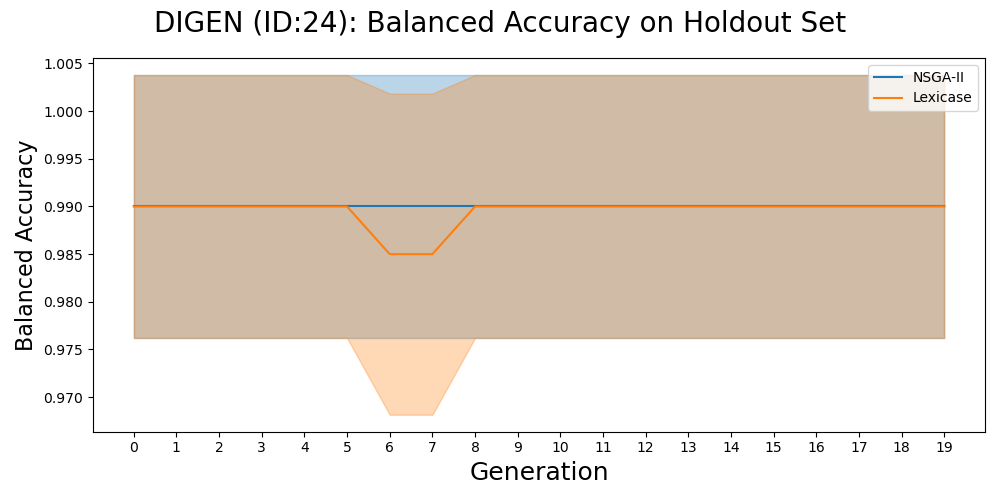}}  \\
    \fbox{\includegraphics[width=5cm]{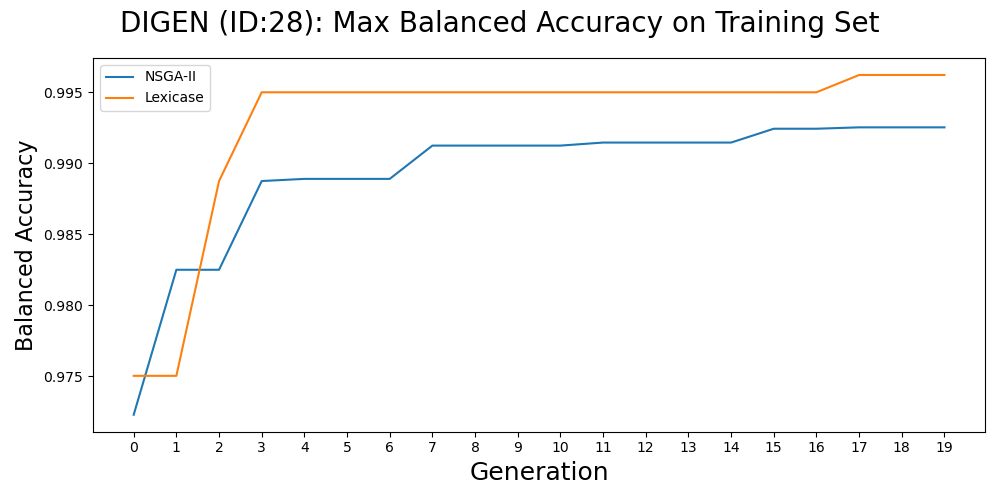}}  & \fbox{\includegraphics[width=5cm]{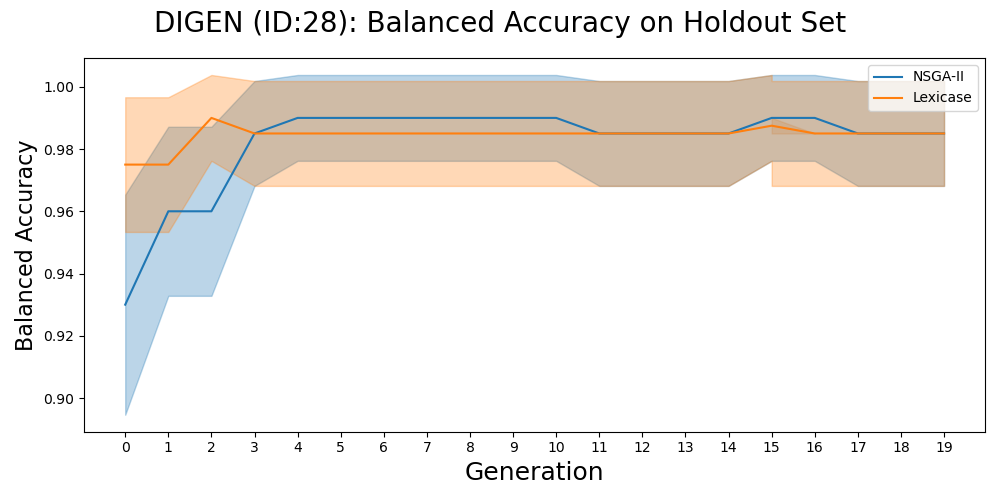}}
\end{tabular}    
   % \centering
    %\includegraphics[width=5cm, height=4cm]{ngsa2.png}
    \caption{Left: Best model based on training accuracy per generation across multiple runs of TPOT for ANGES (50 runs) and select DIGEN datasets (40 runs). Right: Holdout scores with 95\% confidence intervals (+/- 1.96 sd) for the best models based on training accuracy per generation for ANGES and DIGEN.}
    \label{fig:accuracy_training_holdout}
\end{figure}

In Table \ref{tab:convergence}, we show the mean convergence points for various selection methods averaged over 40 runs for each DIGEN dataset and over 50 runs for the ANGES dataset. The procedure to calculate the convergence points has been described in the previous section. To check whether the differences in the convergence points are statistically significant or not, we used a non-parametric test called Mann-Whitney-U test and reported the resulting p-values in the same table. %In Table \ref{tab:holdout_acc}, 
We also report the accuracy of the models on the holdout set found at the convergence points. We show both the mean accuracy across the respective runs and the p-values for the Mann-Whitney-U test applied to the accuracy values on the holdout set. In Figure~\ref{fig:accuracy_training_holdout}, we report the balanced accuracy values of the best model (based on training accuracy) per generation across multiple
runs for ANGES and select DIGEN datasets and for the models with the best accuracy, demonstrate comparable accuracy across methods on the holdout set with 95\% confidence intervals. 

\begin{table}
\centering
  \caption{(a) Mean convergence points for various selection methods on different datasets. The third column shows the p-values obtained by applying the Mann-Whitney-U test on the distribution of values from both selection methods in the corresponding column. (b) Mean accuracy on the holdout set at convergence points for various selection methods on different datasets. The sixth column shows the p-values obtained by applying the Mann-Whitney-U test on the distribution of values from both selection methods in the corresponding columns.}
  \label{tab:convergence}
  \begin{tabular}{c|c|c|c|c|c|c}
    \toprule
     Dataset & NSGA-II (a) & Lexicase (a) & p-value (a)& NSGA-II (b) & Lexicase (b) & p-value (b)\\
     \midrule
ANGES & 16.44	& 10.10 & 8.85E-03 &0.73	&0.73	&0.59\\
DIGEN-2 & 7.65	& 4.70 & 1.26E-18 &0.95	&0.92	&0.06\\
DIGEN-4 &8.97	&5.97 & 3.36E-05 &0.95	&0.95	&0.84\\
DIGEN-7 &10.30	&5.60 & 4.99E-07 &0.97	&0.96	&0.90\\
DIGEN-14 &10.10	&5.60 & 1.86E-06 &0.97	&0.97	&0.85\\
DIGEN-23 &10.32	&6.15 & 2.40E-05 &0.96	&0.94	&0.28\\
DIGEN-24 &4.25	&2.47 & 1.38E-04  &0.94	&0.94	&0.42\\
DIGEN-25 &8.27	&5.37 & 3.66E-05 &0.95	&0.96	&0.31\\
DIGEN-27 &5.42	&2.82 & 4.08E-05 &0.93	&0.93	&0.79\\
DIGEN-28 &8.72	&5.37 & 8.83E-06 &0.93	&0.94	&0.46\\
DIGEN-30 &10.70	&8.95 & 4.79E-02 &0.95	&0.94	&0.34\\
DIGEN-32 &3.35	&3.22 & 6.40E-01 &0.93	&0.94	&0.98\\
DIGEN-35 &5.92	&3.45 & 1.41E-03 &0.94	&0.95	&0.85\\
DIGEN-40 &5.35&	3.42 & 2.73E-03 &0.94	&0.91	&0.31\\
  \bottomrule
\end{tabular} 
\end{table} 

%\vspace*{-1cm}

\subsection{DIGEN Datasets}

For 12 of the 13 DIGEN datasets tested (all excluding DIGEN 32), the selection methods have statistically significant impact on the number of generations used to reach the best cross validation score. On average across all 13 datasets, the convergence point for lexicase is about 2.96 generations sooner. The largest absolute difference was in DIGEN 7 where the average convergence point for lexicase was 4.7 generations sooner than NSGA-II. The selection methods mostly did not have statistically significant differences in maximum balanced accuracy scores on the holdout set. The few datasets that were statistically significant still had differences of less than 0.01. There was also no statistically significant difference in the number of operators used for the best models at the convergence points (Table 1 in supplementary material). Both selection methods ranged between 2 and 4 operators depending on the datasets. As shown in Figure \ref{fig:accuracy_training_holdout}, there does not appear to be any significant difference in the accuracy on the holdout set for lexicase and NSGA-II for DIGEN datasets.

\begin{figure}[h]
     \fbox{\includegraphics[width=6cm]{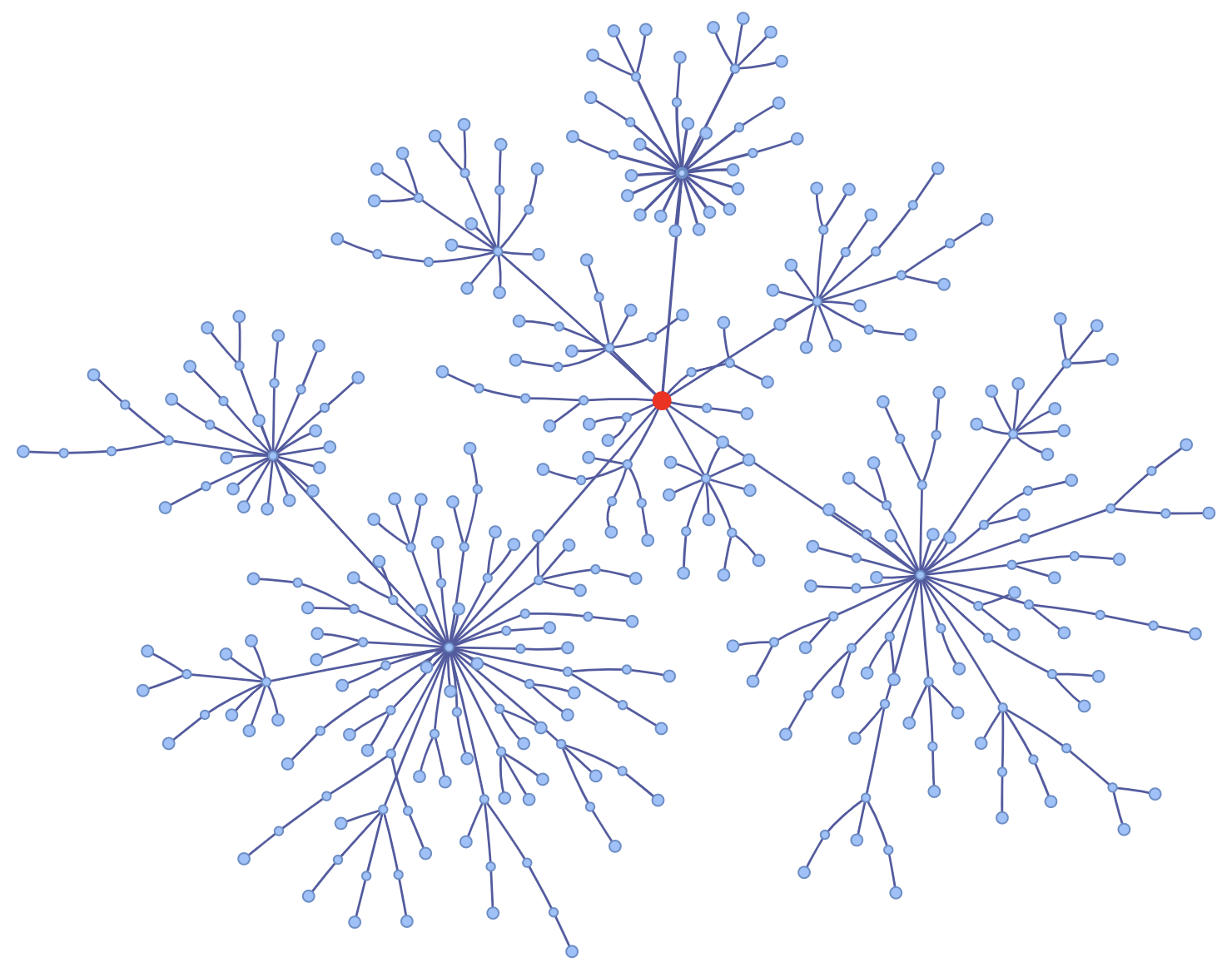}}   
     \fbox{\includegraphics[width=6cm]{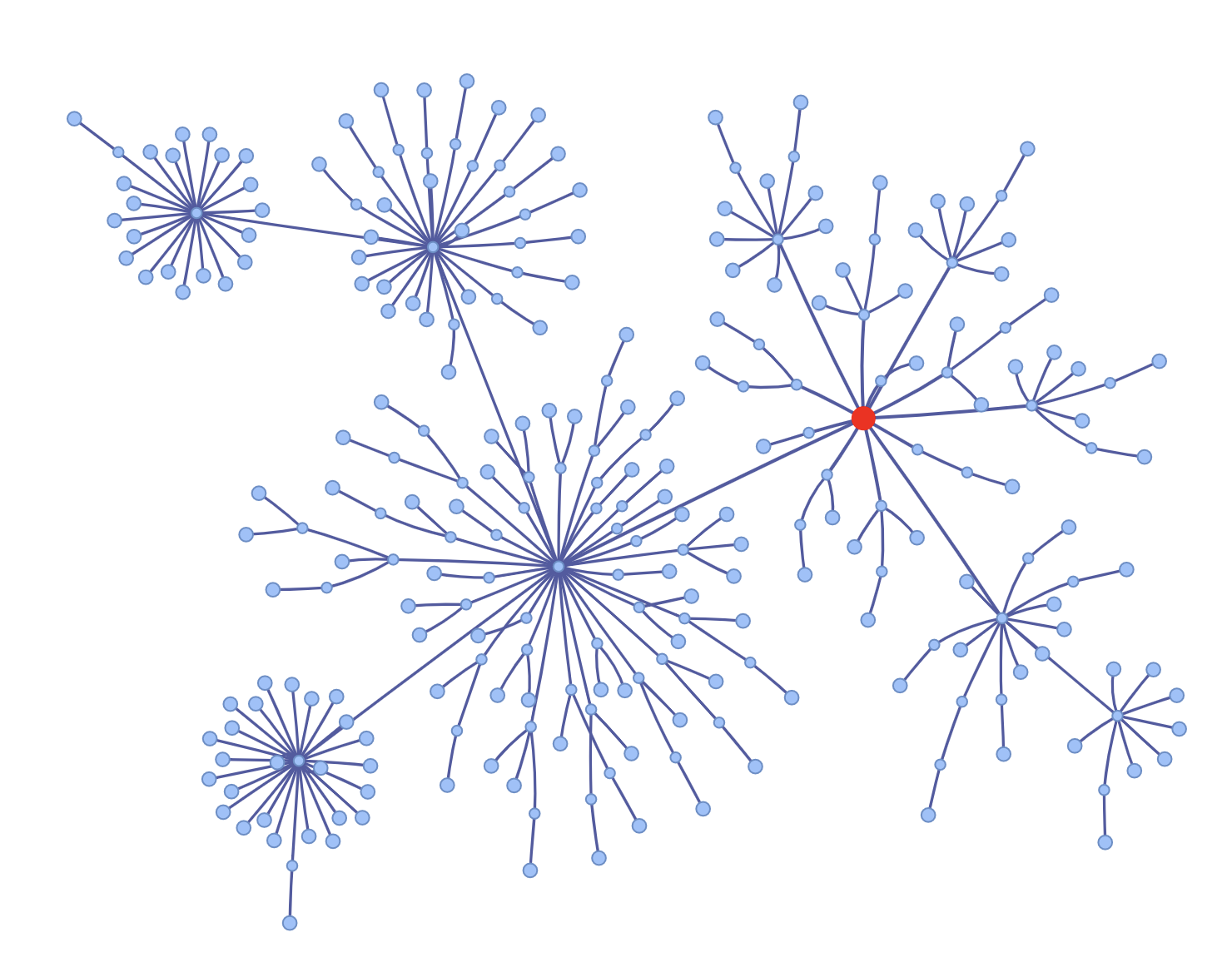}}   
    %\hspace{30px}
    
   % \centering
    %\includegraphics[width=5cm, height=4cm]{ngsa2.png}
    \caption{ Single Run of TPOT on DIGEN-24 dataset using NSGA-II (left), lexicase (right). Red node indicates the root node.}
    \label{fig:digen_trie}
\end{figure}
%\vspace*{-.2cm}

\begin{figure}[h]
     \fbox{\includegraphics[width=6cm]{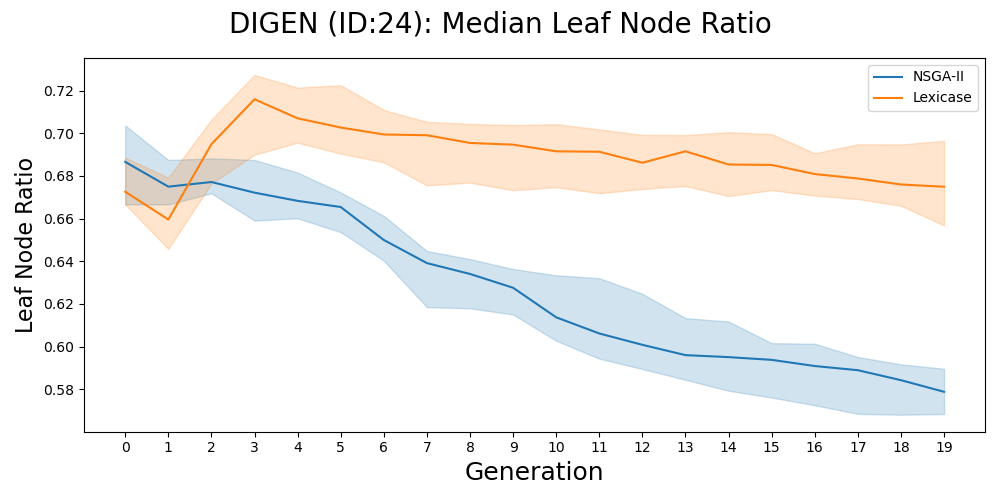}}
     \fbox{\includegraphics[width=6cm]{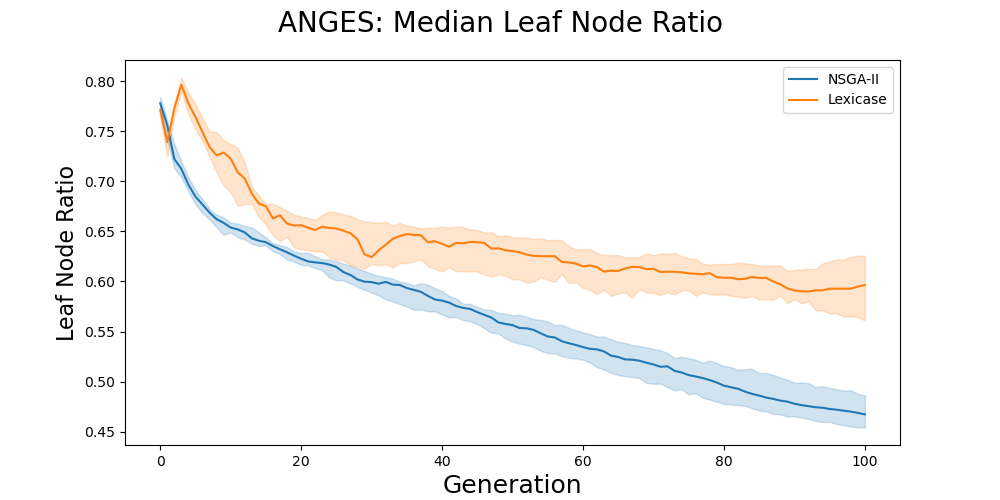}}  
     \fbox{\includegraphics[width=6cm]{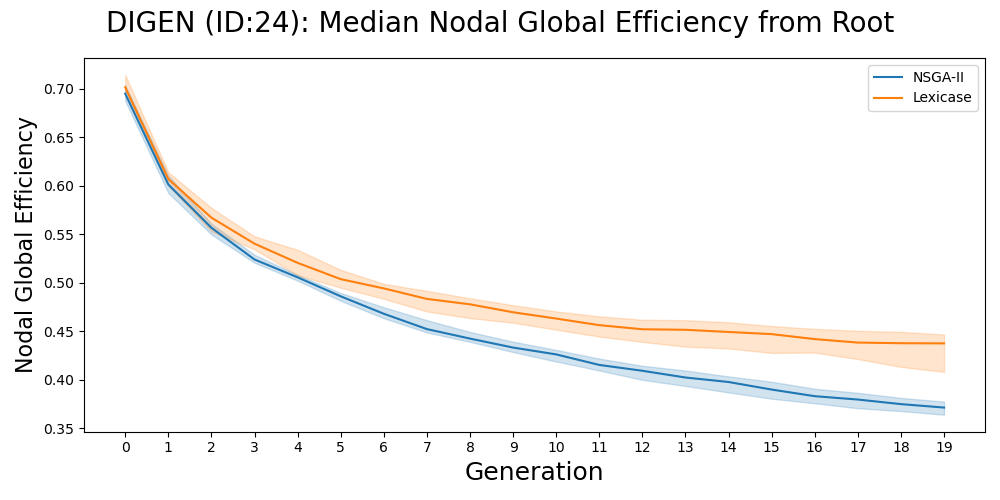}}   
     \fbox{\includegraphics[width=6cm]{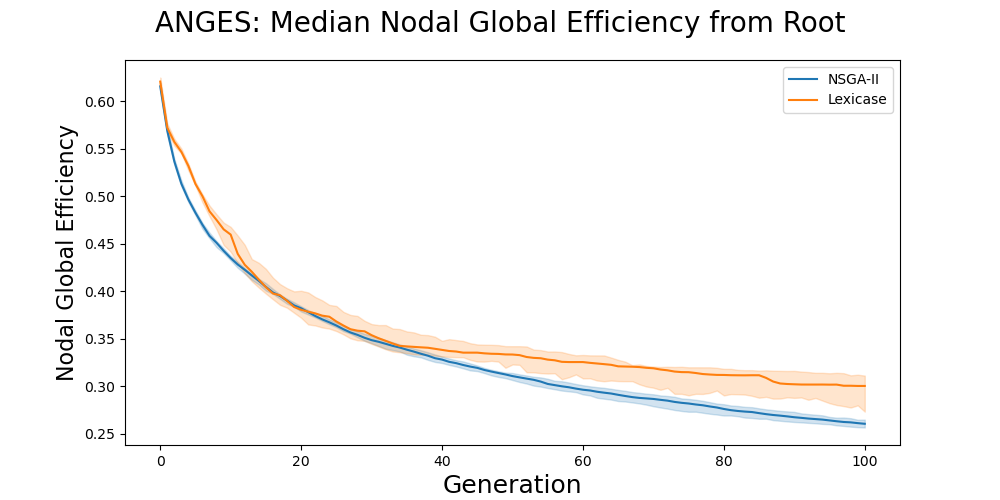}}   
    \fbox{\includegraphics[width=6cm]{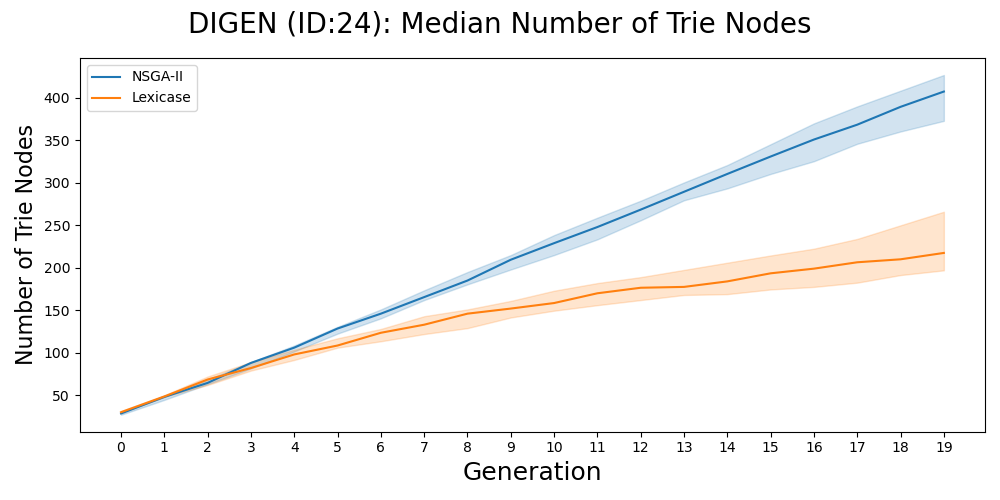}}   \fbox{\includegraphics[width=6cm]{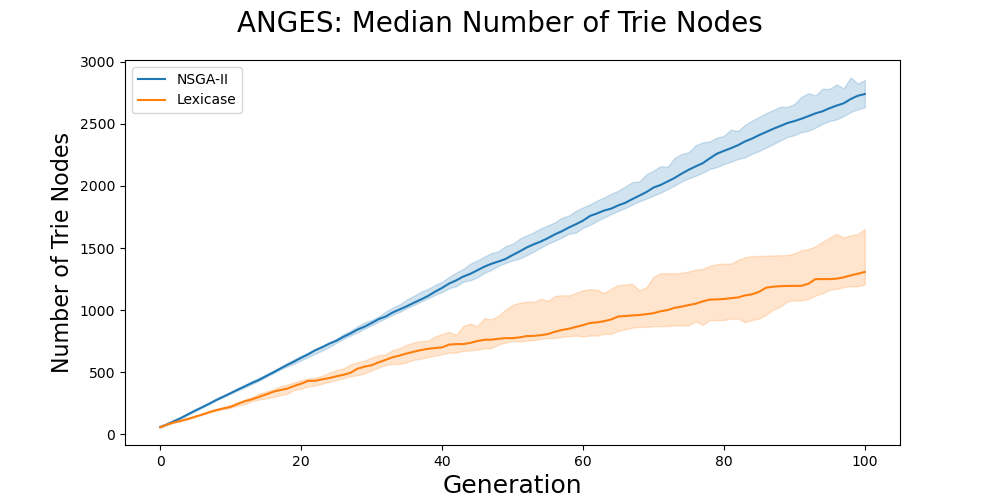}}  
    
   % \centering
    %\includegraphics[width=5cm, height=4cm]{ngsa2.png}
    \caption{Median values of different metrics on TPOT runs on the DIGEN-24 and ANGES datasets. From top to bottom: Leaf-to-node ratio, nodal global efficiency from the root node, total number of nodes in the trie.}
    \label{fig:trie_metrics}
\end{figure}

For all experiments we calculated the median values of leaf-to-node ratio, nodal global efficiency, and the number of trie nodes for TPOT runs and these metrics are summarized in the supplementary material Table 2. As an example in Figure \ref{fig:trie_metrics},  we plot the median values of leaf-to-node ratio, nodal global efficiency, and the number of trie nodes for TPOT runs on the DIGEN-24 and ANGES datasets. The plots for other DIGEN datasets look similar.

When we examine the TPOT tries for the selection variants, NSGA-II explored the largest set of sequences of machine learning operators. To illustrate this with an example in Figure~\ref{fig:digen_trie} and Figure~\ref{fig:trie_metrics} for DIGEN-24, this can be viewed from the number of nodes in each respective graph. The length of sequences of operators observed was also greater for NSGA-II on average than for lexicase. Lexicase had slightly greater root nodal global efficiency compared to NSGA-II (as for example for the DIGEN-24 data set in Figure~\ref{fig:trie_metrics}) indicating shorter sequences of operators on average.  However, the leaf-to-node ratio, the metric of relative branching, drastically differs in the methods. Lexicase maintains the leaf-to-node ratio at nearly 70\% and NSGA-II permits the leaf-to-node ratio to drop to nearly 58\% (as shown in  Figure~\ref{fig:trie_metrics}).

\vspace*{-.4cm}

\subsection{ANGES Datasets}
 On the ANGES dataset the mean convergence point of lexicase of 10.10 generations was significantly shorter than the mean convergence point of NSGA-II at 16.44 generations while there was no statistically significant difference in mean balanced accuracy on the holdout set (as shown in Table~\ref{tab:convergence}) or the number of operators of the best models (Table 1 in supplemental material). When we consider the maximum balanced accuracy on the ANGES training set and the balanced accuracy of the best performing models on the ANGES holdout set in Figure~\ref{fig:accuracy_training_holdout}, both methods reach the same balanced accuracy on the training set and the same balanced accuracy on the holdout set with entirely overlapping confidence intervals at the final generations. 
 
 In terms of the exploration tries on the ANGES dataset in Figure~\ref{fig:trie_metrics}, the median number of nodes in the tries for NSGA-II were more than double the median number of nodes in the tries for lexicase whereas the leaf-to-node ratio for lexicase remained at nearly as 63\% while for NSGA-II it fell to only 48\%. The median root global efficiency of both methods declined to 33\% for lexicase and 25\% for NSGA-II indicating that both methods explores longer sequences of operators for the ANGES dataset. However, the greater leaf-to-node ratio of lexicase and fewer total nodes indicates that even on the ANGES dataset the exploration of the sequences of machine learning operators ends much sooner than in case of NSGA-II. In other words, NSGA-II has the tendency of lengthening sequences of ML operators of many different prefixes, but lexicase usually prefers fewer prefixes and lengthens them.
 
Interpretation of machine learning models is a crucial component of predictive analysis in biomedical studies. So in order to verify that TPOT with both lexicase and NSGA-II are using similar features in their best models, we perform the following analysis. We use permutation feature importance (PFI) analysis that generates the informative coefficients for ANGES models. We perform this analysis on the best models across all the runs based on the accuracy on the training dataset. While PFI can vary based on different initializations, PFI rank shows that most of the top 10 features in both models are the same. Both models selected features in their models that are known well-supported clinical risk factors or predictors for CAD. Detailed information about the top 10 features is given in supplementary material Section 3.

\section{Discussion}
\label{sec:discussion}
When the selection methods are maximizing the balanced accuracy on cross-fold validation and minimizing the number of operators in pipelines, lexicase finds models of comparable balanced accuracy on the holdout set and the number of operators to those found by NSGA-II but in fewer generations. 

To examine the behavior of the selection methods, we introduced the concept of an exploration trie that represents the explored search space of machine learning operator sequences for a TPOT run. Using graph metrics such as nodal global efficiency, leaf-to-node ratio, and total number of nodes in exploration tries, we can further understand the evolutionary selective pressure that lexicase and NSGA-II provide. Observing an overall lower leaf-to-node ratio and higher total number of nodes in exploration tries, NSGA-II explores a larger set of longer machine learning operator sequences compared to lexicase. Lexicase, which has an overall higher leaf-to-node ratio in exploration tries, creates more leaves around an optimal sequence of machine learning operators than is done in the case of NSGA-II.  Comparatively, NSGA-II often keeps the best models for every possible number of operators. This means longer sequences of machine learning operators are considered for pipelines and consequentially longer pipelines are evaluated. Lexicase effectively chooses half its population to have minimal number of nodes while the other half is selected for the best accuracy. It does not automatically keep sequences with a greater number of machine learning operators. 
 
In TPOT, evaluating machine learning pipelines is a resource demanding task. As the sequences of machine learning operators increase and corresponding pipelines include more operators, there are higher computational costs and longer evaluation times. In lexicase the selective pressures of fewer operators in pipelines and highly accurate solutions may lead to the evaluation of fewer pipelines of shorter length, which would in turn reduce the resource costs. NSGA-II, due to its diversity-preserving behavior, is more likely to evaluate a wider variety of pipelines of different lengths, which can often increase resource costs. With lexicase the reduction of average computational resources can permit more allocation of resources for increasing population size or increasing the number of generations if necessary. 

\subsubsection{Acknowledgements} This work is supported by National Institute of Health grants R01 LM010098 and R01 AG066833.

\bibliographystyle{splncs04}
\bibliography{Matsumoto}

\end{document}